\pdfoutput=1

\documentclass[11pt]{article}

\usepackage{EACL2023}

\usepackage{times}
\usepackage{latexsym}

\usepackage{caption}
\usepackage{subfigure}
\usepackage{graphicx}
\usepackage{amsmath}
\usepackage{amssymb}
\usepackage{comment}
\usepackage{multirow}
\usepackage{algorithm}
\usepackage{algpseudocode}
\usepackage{latexsym}
\usepackage{breqn}

\usepackage[T1]{fontenc}

\usepackage[utf8]{inputenc}

\usepackage{microtype}

\usepackage{inconsolata}

\title{Multi-Modal Bias: Introducing a Framework for Stereotypical Bias Assessment beyond Gender and Race in Vision–Language Models}

\author{Sepehr Janghorbani \\
  Rutgers University \\
  New Brunswick, NJ, USA\\
  \texttt{sepehr.janghorbani@rutgers.edu} \\\And
  Gerard de Melo\\
  HPI / University of Potsdam\\
  Potsdam, Germany\\
  \texttt{gdm@demelo.org} \\}

\begin{document}
\maketitle
\begin{abstract}
Recent breakthroughs in self-supervised training have led to a new class of pretrained vision–language models. While there have been investigations of bias in multimodal models, they have mostly focused on gender and racial bias, giving much less attention to other relevant groups, such as minorities with regard to religion, nationality, sexual orientation, or disabilities. This is mainly due to lack of suitable benchmarks for such groups. We seek to address this gap by providing a visual and textual bias benchmark called MMBias, consisting of around 3,800 images and phrases covering 14 population subgroups. We utilize this dataset to assess bias in several prominent self-supervised multimodal models, including CLIP, ALBEF, and ViLT. Our results show that these models demonstrate meaningful bias favoring certain groups. Finally, we introduce a debiasing method designed specifically for such large pretrained models that can be applied as a post-processing step to mitigate bias, while preserving the remaining accuracy of the model.
\end{abstract}

\section{Introduction}

\begin{figure*}[t]
\centering
\includegraphics[trim={0 6cm 0 0},clip,scale=0.22]{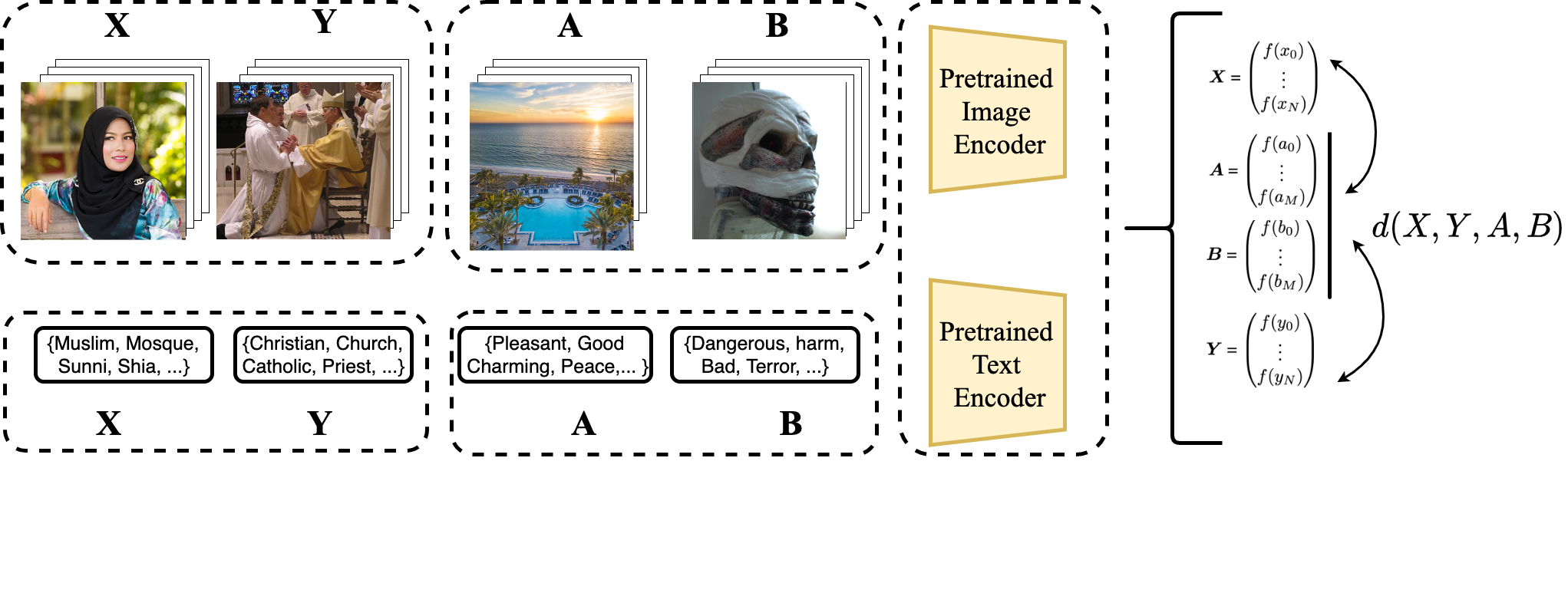}
\caption{Experiment pipeline. We feed target and attribute data to the model. The embeddings obtained from pretrained encoders are then used to measure bias metrics between visual and textual stimuli following Eq.~\ref{eq:bias_def}.}
\label{fig:pipeline}
\end{figure*}

The recent emergence of large pretrained vision--language models has revolutionized many multimodal tasks previously considered impractical to solve. 
Although architectures capable of jointly addressing computer vision and NLP tasks using a single unified model have been around for a while \cite{lu2019vilbert,tan2019lxmert,li2019visualbert}, recent advances in self-supervised training methods have amplified the significance and applicability of such models. The sheer power of these methods is highly dependent on the scale of the model and the diversity and distributional properties of the dataset on which they are trained. Due to their wide range of diverse applications \cite{PubMedCLIP}, it is of utmost importance to be aware of the shortcomings of vision--language pretrained (VLP) models as well as their capabilities.

One such limitation is that, like any other machine learning system, multimodal models may be prone to exhibiting human-like stereotypical biases such as gender or race-related stereotypes \cite{nadeem2020stereoset,garrido2021survey}. For instance, pretrained language models have been shown to associate male-gendered phrases and sentences to a greater extent with certain high-paying professions and even with individual traits such as intelligence, in comparison to female-gendered phrases \cite{wang2021gender}. Similarly, it has been found that Hispanic and African American names may be tied to words representing danger and crime more often than Caucasian names \cite{manzini2019black}. Certain biases have also been identified in computer vision models as well \cite{wang2019balanced}.
Such biases are discriminatory towards affected population groups and can be extremely harmful to society the more these models are deployed in real-world applications.

While there has been some research aimed at identifying and addressing biases in vision--language models, most such studies have focused on gender and racial biases, while other notable groups such as religious minorities, national minorities, LGBTQ people, and people with disabilities have received much less attention, despite their legal status as protected groups in the US. 
This is alarming considering the fact that the potentially affected groups together constitute a considerable part of the global population. For instance, the US Census Bureau reported approximately 40 million people identifying as immigrants in the US and 244 million world-wide as of 2015.\footnote{www.un.org/en/development/desa/population/migration} Furthermore, approximately 40 million people in the US and about 1 billion people in the world suffer from some sort of disability.\footnote{www.worldbank.org/en/topic/disability}
One of the main obstacles for bias analysis of these relevant population groups has been the lack of standardized benchmark datasets that specifically enable an analysis of how they may be affected.
In this paper, we attempt to address this problem by gathering and releasing a visual and textual bias benchmark called MMBias, consisting of approximately 3,500 images and 350 phrases covering over 14 minority subgroups. Furthermore, we utilize the dataset to measure stereotypical bias in several prominent self-supervised multimodal VLP models that have attracted significant attention recently, namely Open\-AI CLIP \cite{radford2021learning}, ALBEF \cite{li2021align}, and ViLT \cite{kim2021vilt}. In our experiments, we quantify the bias present in these models, including both cross-modal and intra-modal bias.
Our results confirm that these models harbor meaningful biases favoring certain groups. Finally, we introduce a novel debiasing method designed for such large pretrained models that can be applied as a post-processing step to mitigate bias, and we show that this step does not adversely affect the performance in a substantial way.

\begin{figure*}[t!]
\includegraphics[width=\textwidth, height=0.3\textwidth]{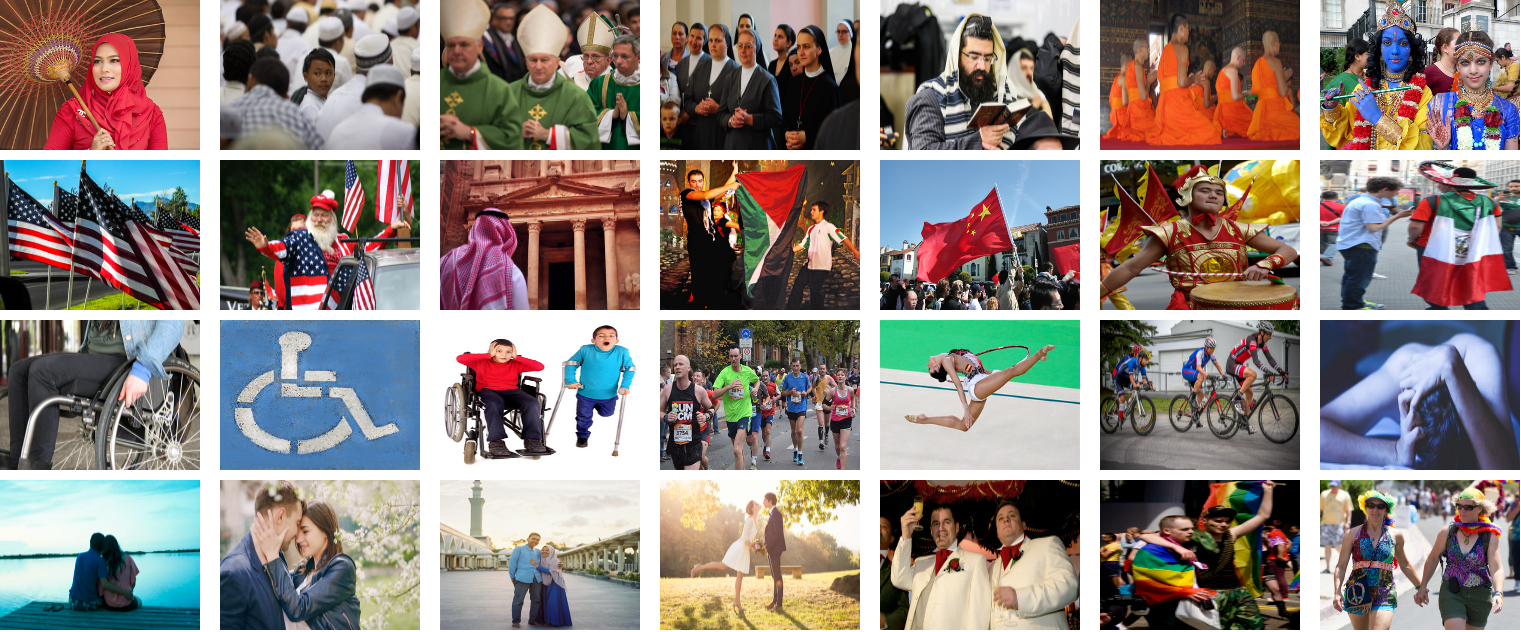}
\caption{Sample images from the MMBias dataset. Each row corresponds to one of the target classes: religion, nationality, disability, and sexual orientation. Images are compiled from the image sharing service Flickr.}
\label{fig:samples}
\end{figure*}

\section{Related Work}

The majority of work on language models only focuses on gender and racial bias assessment \cite{guo2021detecting,bordia2019identifying}. However, there have been some studies that consider bias with regard to other categories such as profession, religion, and disability as well \cite{nadeem2020stereoset,hutchinson2020social}. 
However, these forms of bias are not just exclusive to the language domain, and image classifiers as well as multimodal models have also been shown to demonstrate such biases \cite{srinivasan2021worst,ross2020measuring}.

Thus far, there has been rather limited work on multimodal bias assessment of self-supervised models such as CLIP, and prior work considers only gender and racial biases.
\newcite{wang2021assessing} measures the gender and racial bias in CLIP's image classification module using the Fairface dataset, while \newcite{wang2021gender} further show that CLIP associates male-gendered phrases to high-paying professions more than female-gendered phrases. 
\newcite{agarwal2021evaluating} provide insights towards potential applications of the CLIP model and further study and evaluate its gender/racial bias as well as measuring the misclassification differences between different subgroups.
\newcite{bhargava2019exposing} measure and propose solutions to gender bias in several image captioning systems. 

The only work that addresses other relevant groups such as religion, sexual orientation, and disability in the image space is \newcite{steed2021image}. However, the data considered in the study is limited, consisting of only around 600 images, out of which around 500 again correspond to gender and racial biases. This leaves only around 100 for other protected groups, i.e., fewer than 20 images for each protected group study. 
\newcite{sirotkin2022study} use this limited dataset to measure bias in several self-supervised visual models but the authors do not explore multimodal models such as CLIP. An orthogonal line of research has been pursued in \newcite{zhou2022vlstereoset}, where several multimodal vision-language models are  analyzed to measure these models' tendency
to pick stereotypical statements as captions
for anti-stereotypical images in pre-trained vision-language models.
With MMBias, we thus hope to enable further research on diverse forms of bias in vision--language models.

\section{Methodology}

\paragraph{Bias and Fairness.}
In conventional social studies, one of the most well-established and widely-investigated forms of biases is what is known as implicit bias, or as social stereotypes, defined and investigated in \citet{greenwald1995implicit}. This type of bias is usually measured using Implicit Associate Testing (IAT), introduced in the seminal work of \cite{greenwald1998measuring}, and has so far been widely used  to describe and account for a wide range of implicit prejudices \cite{kiefer2007implicit}. IAT experiments quantify human implicit bias by measuring response times differences when human subjects are asked to pair similar concepts and different concepts. In its original form, IAT was used to to measure the degree of \emph{pleasantness} (a.k.a.\ \emph{valence} in psychology), of entities such as ``flowers'' and ``insects'' by pairing them with abstract attributes such as \emph{pleasant} and \emph{unpleasant} \cite{russell2003core}. \citet{caliskan2017semantics} showed that a similar IAT testing paradigm can be applied to bias measurement in deep embeddings. In this approach, instead of subject reaction time, the proximity of embeddings of a basket of words that represent a concept is measured. Furthermore, word sentiment is usually used to represent valence, due to well-established studies linking word sentiment with the psychological concept of valence \cite{mohammad2016sentiment}. The experimental methodology used in our study follows similar principles.

More generally, a machine learning system may be deemed \emph{unbiased} or \emph{fair} when its predictions do not favor members of any relevant population group or discriminate against any other \cite{garrido2021survey}. For instance, suppose that the class under consideration is religion and we are evaluating \emph{pleasantness} / \emph{unpleasantness} scores a system would assign to each considered religious subgroup. A machine learning system is assessed as fair if and only if the scores it assigns to different religious subgroups do not differ substantially.

More formally, in a bias study, the two subgroups under study, also known as target entities, may be represented as sets of instances $\boldsymbol{X}=\{x_1,x_2,...,x_N\}$ and $\boldsymbol{Y} =\{y_1,y_2,...,y_N\}$. For example, $\boldsymbol{X}$ may be images corresponding to Islam and $\boldsymbol{Y}$ to Christianity. Furthermore, the attributes towards which the bias is being measured may be given as sets $\boldsymbol{A} =\{a_1,a_2,..., a_M\}$ and $\boldsymbol{B} =\{b_1,b_2,..., b_M\}$. For example, $\boldsymbol{A}$ could be a set of words  representing \emph{pleasantness}, while $\boldsymbol{B}$ represents \emph{unpleasantness}. 

Similarly, many gender-bias studies consider sets for \emph{high paying} vs.\ \emph{low paying} professions as attribute sets. A machine learning model is then said to be fair towards subgroups $\boldsymbol{X}$ and $\boldsymbol{Y}$ with respect to attributes $\boldsymbol{A}$ and $\boldsymbol{B}$ if and only if $\phi(\boldsymbol{X}, \boldsymbol{A}, \boldsymbol{B}) \approx \phi(\boldsymbol{Y}, \boldsymbol{A}, \boldsymbol{B}) $, where $\phi$ is some scoring function that scores the similarity of the sets of attributes $\boldsymbol{A}$, $\boldsymbol{B}$ to a target entity $\boldsymbol{X}$ or $\boldsymbol{Y}$.

\paragraph{Scoring Functions.}
There can be different choices for the scoring function $\phi$ above.
\newcite{caliskan2017semantics} introduced the \emph{Caliskan} test shown below in Eq.~\ref{eq:bias_def}, with $\phi$ capturing the difference of the mean of cosine distances between targets and attributes. This method is ideal for the analysis of models such as CLIP, since they operate directly on entity embeddings. The effect size represented by $d(X,Y,A, B)$ is a measure of the magnitude of the bias. Larger numbers indicate a stronger bias, while the sign reflects which target entity the attributes show a stronger bias towards. 

However, for vision--language fusion models that do not provide explicit access to separate image/text embeddings, an alternative scoring function can be defined as the difference in the image--text matching probabilities, as in the last row of Eq.~\ref{eq:bias_def}. Sets $\boldsymbol{X}$ and $\boldsymbol{Y}$ as well as $\boldsymbol{A}$ and $\boldsymbol{B}$ are usually constructed to have equal number of samples.

\vspace{-3mm}
\begin{equation}\label{eq:bias_def}
d=\frac{\underset{x \in X}{\text{mean}} \ \phi(x,A,B) - \underset{y \in Y}{\text{mean}} \ \phi(y,A,B)}{\underset{w \in X \cup Y}{\text{std-dev}} \ \phi(w,A,B)}
\end{equation}
\vspace{-3mm}

\begin{align*}
\phi(w, A, B) &=
\underset{a \in A}{\text{mean}} \ \cos(w, a) - \underset{b \in B}{\text{mean}} \ \cos(w, b) 
\\
\phi(w, A, B) &=
\underset{a \in A}{\text{mean}} \ \sigma(w, a) - \underset{b \in B}{\text{mean}} \ \sigma(w, b)
\end{align*}

\noindent Here, $\cos(\cdot,\cdot)$ denotes the cosine similarity of vectors, while $\sigma(\cdot,\cdot)$ denotes the probability of a text and image pair being a match.

\paragraph{Evaluation Pipeline}
Fig.~\ref{fig:pipeline} shows the pipeline followed in our experiments. The target and attribute stimuli are fed into the model and embeddings it emits are used to compute the bias score.

\section{The MMBias Dataset}
\label{sec:dataset}
The majority of the work on societal bias analysis so far focuses on unimodal language models. Although there has been some limited work on multimodal models, these studies mainly focus on gender and racial disparities. As a result, bias with regard to other classes, including religion, nationality, sexual orientation, and disability have largely been unexplored. This has been mainly due to the lack of standardized benchmark datasets that specifically target these minority groups. To address this concern, we gather and release the first multimodal dataset of this size in this line of research that spans over a wider range of groups. We hope that this dataset can serve as a benchmark in future research.

\begin{table}[!b]
\centering
\resizebox{\linewidth}{!}{
\begin{tabular}{ |c|c| }
\hline
Target Concept $\boldsymbol{X}$ & Target Values $\{x_1,...,x_N\}$ \\
\hline

\multirow{2}{*}{Religion} & Islam, Christianity, Judaism,\\
& Buddhism, Hinduism \\
\hline
 \multirow{2}{*}{Nationality} & American, Arab,\\
& Chinese, Mexican \\ 
\hline
 \multirow{2}{*}{Disability} & Physical disability, \\
 & Mental disability, No disability \\
\hline
Sexual Orientation & Homosexual, Heterosexual \\ 
\hline
\end{tabular}
}
\caption{MMBias spans over 4 target classes and 14 target groups including 5 major religions, 4 nationalities, 2 forms of disability and sexual orientations.}\vspace*{1mm}
\label{tab:targets}
\end{table}

\begin{figure}[t!]
\includegraphics[scale=0.3,trim={30 0 0 60},clip]{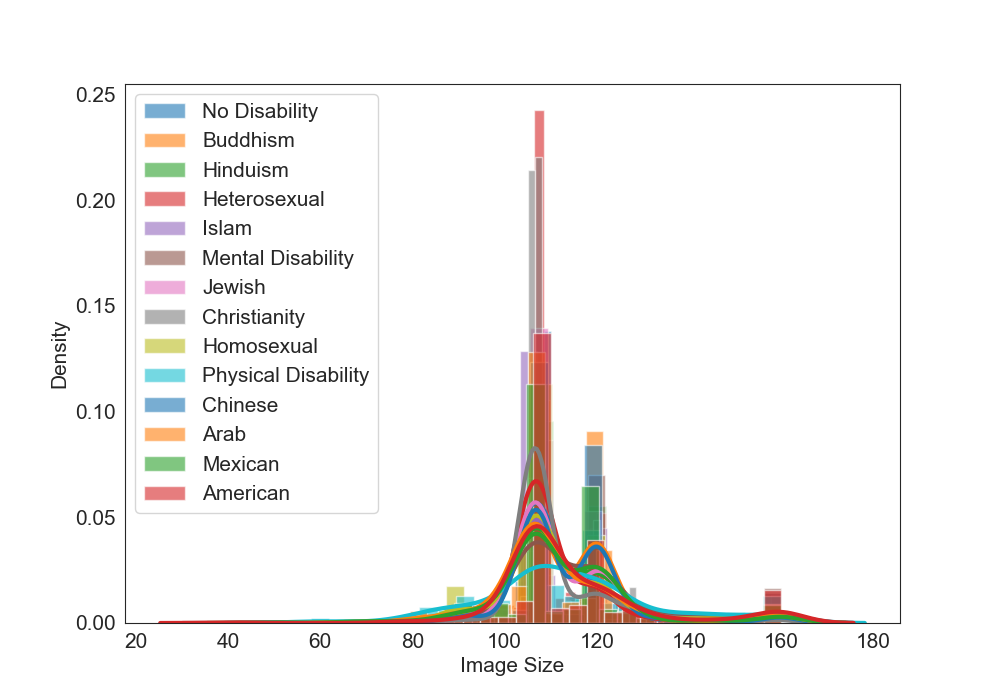}
\caption{Gaussian distribution of the image sizes scaled by a factor of 1,000. Most images are sized around 340x340 pixels.}
\label{fig:sizes}
\end{figure}

Our dataset, referred to as \textbf{MMBias}, contains 3,500 target images and 350 English phrases corresponding to different target concepts. Each target category has 250 corresponding images obtained from the popular image uploading website Flickr. Our dataset also contains 20 textual phrases related to each target concept, used for bias experiments in the textual domain.

\paragraph{Data Compilation}
For gathering the image data, we invoke the Flickr API and retrieve 1,000 most relevant images for each target concept using 10--12 search keywords for each. The keywords are chosen to be as diverse as possible to minimize any potential bias in the data gathering process to the extent possible. Then human annotators are used to filter out noisy images. Annotators were instructed to manually eliminate irrelevant or explicit content as well as images that contained private information or names/addresses. In order to balance the dataset, 250 images were randomly chosen for each concept out of the filtered images, and we only consider images with a Creative Commons license. The processing pipeline and quality control is similar to the one used for the creation of Flickr30k \cite{young2014image}. Furthermore, the textual part of the dataset contains phrases such as ``This is \textit{X}.'' replacing \textit{X} with a ``Muslim person", ``Christian person", etc. The same aforementioned keywords were used to retrieve textual data for each concept using the RelatedWords site\footnote{relatedwords.org/} followed by a similar data cleaning and noise filtering process.

Table~\ref{tab:targets} shows the classes MMBias covers as well as the considered groups in each class. 
MMBias spans over 4 target classes, including religion, national origin, disability, and sexual orientation. In this study, we did not include gender and race, as there is already a large body of work focusing on them. 

For religion, our dataset includes the 5 major religions in the world today: Islam, Christianity, Judaism, Buddhism, and Hinduism. As for the national origin, MMBias includes images corresponding to the four nationalities: American (USA), Chinese, Arab\footnote{\emph{Arab} collectively refers to a number of Arab countries (each also having other cultural groups).
We hope that more specific nationalities and cultural groups can be added in the future.} (collectively), and Mexican. Furthermore, another two additional nationalities, French and Italian, are also included in the textual phrases in addition to the former. As for disability, MMBias contains images for two common forms of disability: physical disability, mental disability as well as people with no disability. In addition to these, the textual data includes phrases corresponding to visual disability and hearing disability as well. Finally, the two most common types of sexual orientations, homosexuality and heterosexuality, are included in MMBias. The selection of subgroups as well as their pairings was a result of consulting several social studies that show present bias against
people with disabilities \cite{dovidio2011implicit}, homosexuals \cite{hebl2002formal}, certain nationalities
\cite{park2007implicit,buriel1982stereotypes} and religions \cite{abid2021large,rowatt2005patterns,rudman2007discrimination}. However, we plan to extend our data to a larger pool of classes and respective subgroups in the future.

Finally, in order to conduct intra-visual bias studies, MMBias also contains two sets of images corresponding to visual pleasantness and unpleasantness, called the valence dataset. These sets were constructed by following a similar method to \citet{steed2021image}, retrieving images corresponding to pleasant concepts such wealth, peace, babies, love, butterflies, etc.\ and unpleasant concepts such as death, injury, prison, fear, etc.

\paragraph{Analysis.}
Fig.~\ref{fig:samples} provides some sample images taken from the dataset. Each row shows a different target class. Fig.~\ref{fig:sizes} provides deeper insights into the sizes of the crawled images. The x-axis reflects the surface area of the image in pixels, scaled by a factor of 1,000. As can be seen, image sizes in most classes follow a normal distribution with a mean of around 110,000 pixels, translating to approximately 340x340 images, with the exception of images corresponding to the nationality class, which have a slightly higher mean of around 350x350. The height and width of images does not vary substantially across the dataset. 

\begin{figure}[t!]
\centering
\includegraphics[scale=0.11]{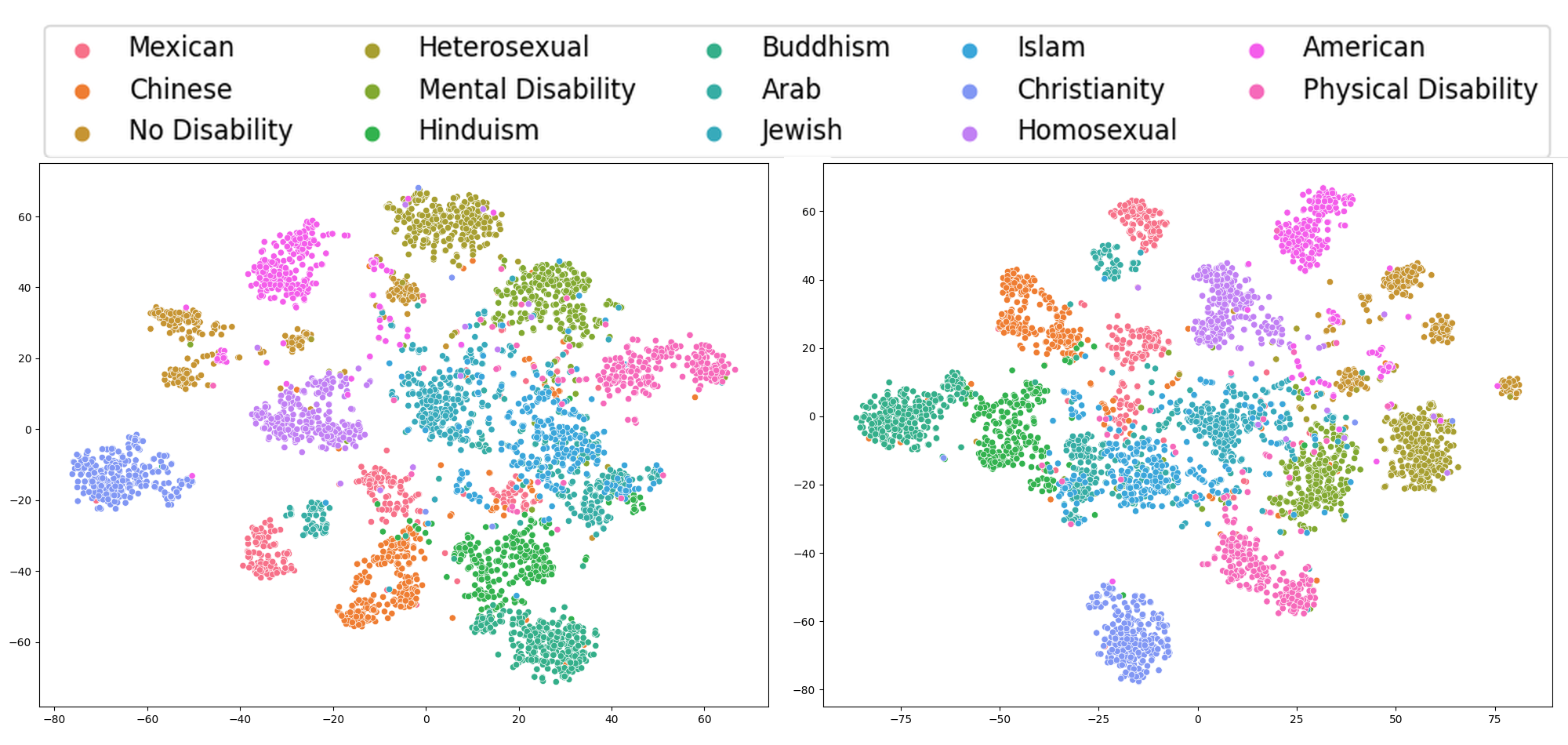}
\caption{t-SNE representation of image embeddings. Left shows embedding clusters before bias mitigation. Right shows  embedding clusters after bias mitigation. Both cases show well-separated clusters, suggesting bias mitigation has negligible effects on cluster separability.}
\label{fig:tsne}
\end{figure}

Furthermore, we analyze the separability of our dataset with regards to image classes. The images are fed into the CLIP model and the first 100 principal components are extracted from the resulting embeddings, and then t-SNE is applied.
Fig.~\ref{fig:tsne} shows the t-SNE representation of the images. We observe that the dataset can be well-separated forming clearly-defined clusters.
For instance, we notice that different religions form well-separated clusters. Interestingly the clusters that are more intertwined correspond to correlated subjects such as the religion Islam and the Arab nationality designation. This is not surprising given Islam is particularly prevalent in Arab countries and thus many of the images share similar features.

\begin{figure*}[t!]
\centering

\includegraphics[width=\linewidth,trim={0 0 0 0},clip]{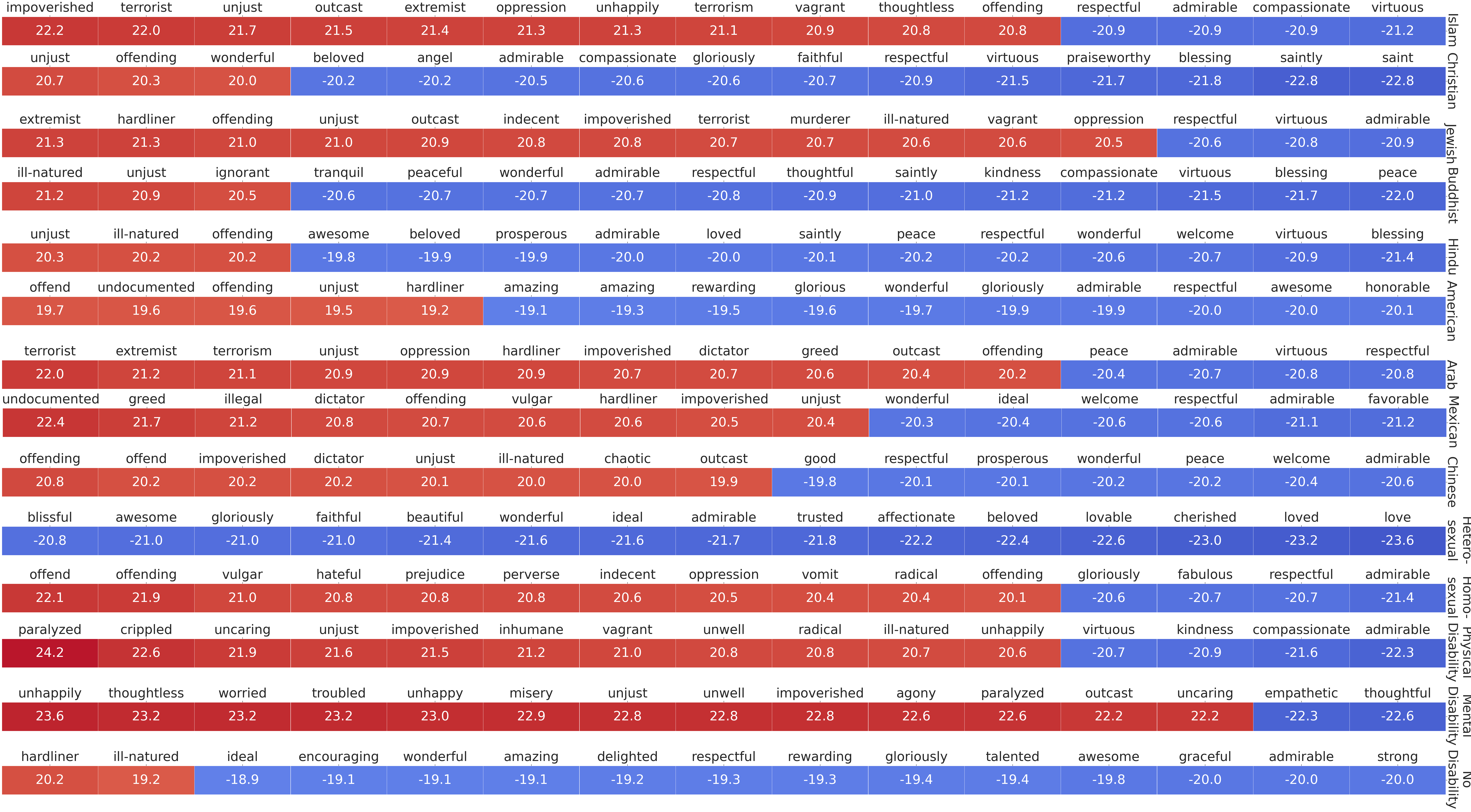}

\caption{Top 15 closest attributes returned by CLIP for each target group. Red colors represent negative sentiment while blue
represents positive sentiment. Stereotypical patterns can be seen among different groups.}
\label{fig:top15} 
\end{figure*}

\section{Experimental Evaluation}
We have conducted three sets of experiments to assess and quantify the bias in the aforementioned models: CLIP, ALBEF, and ViLT. The following sections explain the details of each setting. 

\subsection{OpenAI CLIP}
CLIP is a multimodal vision--language embedding model originally devised for zero-shot classification of images. It utilizes a self-supervised contrastive loss to learn a joint embedding space for both images and text. 
The model is jointly trained on the WebImageText dataset, a set of 400 million paired image--text pairs crawled form the web.
Although primarily designed for image classification, CLIP embeddings have been used in numerous other downstream applications, making it a prime candidate for our analysis. The architecture of CLIP has independent visual and textual encoding modules  providing explicit access to each modality's embedding. As a result, it is possible to not only analyze the bias across domains but conduct ablation studies for each module separately as well. We used the ``ViT-B/32'' model with the official CLIP code. Our experiments are as follows.

\subsubsection{Cross-Modal Zero-Shot Classification}
CLIP was originally introduced as a means for zero-shot image classification. In this experiment, we measure bias for this task across different modalities. Given a set of target concept images $\boldsymbol{X}^I$ and $\boldsymbol{Y}^I$, and a set of textual attribute phrases $\boldsymbol{A}^T$ and $\boldsymbol{B}^T$, we use CLIP to perform zero-shot classification of target images to attribute words. 
For each image group, the top 15 classified words are returned. The attributes are two sets of 60 words conveying positive\footnote{ptrckprry.com/course/ssd/data/positive-words.txt} or negative\footnote{ptrckprry.com/course/ssd/data/negative-words.txt} sentiment, many of which were also included in the
original IAT studies \cite{bellezza1986words}.

The resulting correlation scores are provided in Fig.~\ref{fig:top15}. Each row shows the top 15 words returned by the model for each of the target classes. Words with positive sentiment are blue-colored while adjectives with negative sentiment are given in red. The number inside each bar as well as the color intensity represent the degree to which the model associates that target class with that word. 
Fig.~\ref{fig:top15} shows stereotypical patterns emerging, e.g., the most associated attributes to Islam and Judaism are words related to poverty, terror, and extremism such as: ``impoverished'', ``vagrant'', ''terrorist``, ''oppression``, ``outcast'', ``extremist'', etc., which carry a highly negative sentiment.

However, unlike Islam and Judaism, in the case of Buddhism and Christianity, 12 of the 15 top attributes have positive sentiment. The most associated attributes are words resembling peace and happiness such as: ``peace'', ``blessing'', ``compassionate'', ``admirable'', etc., carrying a highly positive sentiment.
This is aligned with societal stereotypes that certain religions are looked upon less favorably than others.
Similar patterns can be observed for other classes such as nationality as well. Certain nationalities such as Americans are viewed as more favorably by the model compared to Arab, Mexican, and Chinese people. Interestingly, biases against the Arab category are very similar to biases against Islam, e.g., both obtaining high scores for ``terrorism'', ``extremist'', and ``impoverished''. This likely stems from the the fact that most Arab countries are predominantly Muslim and the model may have acquired latent correlations among the two.
For the target class ``Mexican'', the highest-scoring words are ``undocumented'', ``greed'', and ``illegal'', followed by ``impoverished'', which reflects the typical right-wing media portrayal of this group in the US. Similarly, Chinese nationals are associated with negative attributes relating to poverty and dictatorship. The next category that exhibits a bias is sexual orientation, where the LGBTQ community is mostly associated with words such as ``offending'',``vulgar'', ``hateful'', ``perverse'', etc.
Finally, we can see the large negative sentiment the model exhibits towards people with disabilities.

\newcommand{\Ximage}{\boldsymbol{X^{\rm I}}}
\newcommand{\Yimage}{\boldsymbol{Y^{\rm I}}}

\subsubsection{Cross-Modal Bias Assessment} 
This experiment quantifies the bias in CLIP using Caliskan cosine similarity metric in Eq.~\ref{eq:bias_def}. Given a set of target concept images $\Ximage$ and $\Yimage$ and a set of textual attribute phrases $\boldsymbol{A^T}$ and $\boldsymbol{B^T}$, the goal is to measure the effect size, $d(\boldsymbol{X^I},\boldsymbol{Y^I},\boldsymbol{A^T}, \boldsymbol{B^T})$, distance between image target concepts $\boldsymbol{X^I}$, $\boldsymbol{Y^I}$ and textual pleasantness attributes. The results are provided in the first column of Table~\ref{tab:bias}.
Positive numbers reflect a negative bias towards the first target $\boldsymbol{X}$, while negative numbers indicate a positive bias towards $\boldsymbol{X}$. The magnitude represents the intensity to which the bias is present in the model with regard to test data. The results in Table~\ref{tab:bias} are consistent with the results in the zero-shot classification experiment, confirming certain societal stereotypes. For instance, in the case of religion, we have already observed that Islam and Judaism are tied to negative words much more frequently, compared to Christianity and Buddhism. Similarly, here, we observe that bias scores for ``Islam vs.\ Christianity'' and ``Judaism vs.\  Christianity'' are fairly high as well. In the case of ``Islam vs.\  Judaism'', Islam is viewed more unfavorably, reflecting the surge of Islamophobic tendencies in recent decades. In this regard, the most favorably assessed religions are Christianity and Buddhism.
Similar trends can be seen when considering nationality as well. The model shows a negative bias towards Arab, Chinese, and Mexican people compared to Americans. This is again consistent with previous observations in the zero-shot classification experiment. Finally, we find that people with disabilities as well as the LGBTQ community are viewed more negatively.

\begin{table}[!t]
\centering
\scalebox{0.65}{
\begin{tabular}{|c|c|c||c|c|c||c|c|}
 \hline
 & Target & Target & CLIP & CLIP & CLIP & ALBEF & ViLT \\
 & $\boldsymbol{X}$ & $\boldsymbol{Y}$ & Cross & Textual & Visual & &\\
  \hline
\multirow{4}{*}{\rotatebox{90}{Religion \hspace{1cm}}} 
& Muslim & Christian & 1.72 & 1.48 & 1.61 & 0.37 & 0.45\\
& Jewish & Christian & 1.69 & 1.24 & 1.43 & 0.34 & 0.51\\
& Muslim & Jewish & 0.47 & 0.41 & 0.75 & 0.03 & -0.04\\
& Muslim & Buddhist & 1.62 & 0.69 & 1.53 & 0.23 & 0.26\\
& Christian & Buddhist & -0.75 & -0.99 & -0.35 & -0.14 & -0.21\\
& Hindu & Buddhist & -0.52 & 0.06 & -0.11 & 0.01 & 0.01\\
& Jewish & Buddhist & 1.61 & 0.31 & 1.28 & 0.20 & 0.30\\
& Muslim & Hindu & 1.65 & 0.64 & 1.49 & 0.24 & 0.25\\
\hline
\multirow{4}{*}{\rotatebox{90}{Nationality \hspace{0.7cm}}} 
& Arab & American & 1.28 & 1.79 & 1.56 & 0.11 & -0.03\\
& Arab & French & -- & 1.79 & -- & -- &--\\
& Arab & Italian & -- & 1.25 & -- & -- &--\\
& Mexican & Arab & -0.32 & 0.24 & -0.92 & -0.04 &0.06\\
& Chinese & American & 0.89 & 1.30 & 1.20 & 0.03 &-0.07\\
& Mexican & American &  1.13 & 1.75 & 1.20 & 0.07 &0.03\\
\hline
 \multirow{2}{*}{\rotatebox{90}{Disability\hspace{2mm}}} 
 & Visual & Abled & -- & 1.25 & -- & -- &--\\
 &  Hearing  & Abled & -- & 1.13 & -- & -- &--\\
 &  Mental & Abled & 1.48 & 1.04 & 1.05 & 0.37 & 0.13\\
 &  Physical & Abled & 1.25 & 1.03 & 1.35 & 0.02 &-0.01\\
 \hline
 \rotatebox{90}{LGBTQ \hspace{1mm}}
 & LGBTQ & Hetero. & 1.67 & 0.93 & 1.46 & 0.07 & 0.10\\

  \hline
\end{tabular}
}
\caption{Bias assessment for CLIP, ALBEF, and ViLT models. CLIP-Cross has numbers for cross-modal bias assessment experiment, while CLIP-Textual and CLIP-Visual show effect sizes for intra-modality ablation studies. Positive numbers favor target $Y$ while negative numbers favor target $X$.}
\label{tab:bias}
\end{table} 

\subsubsection{Ablation: Intra-Modal Encoder Bias}
Since CLIP provides explicit access to textual and visual embeddings, we can run ablation studies by measuring the bias in each module independently. In order to do so, we measure the  effect size using the Caliskan formula $d(\boldsymbol{X^T},\boldsymbol{Y^T},\boldsymbol{A^T}, \boldsymbol{B^T})$ for textual data and Image Association test $d(\boldsymbol{X^I},\boldsymbol{Y^I},\boldsymbol{A^I}, \boldsymbol{B^I})$ for images, where unlike the cross-modal experiment, both target concepts as well as attributes have the same modality. These experiments can provide insights as to which module is more heavily responsible for the observed bias. Columns 2 and 3 in Table~\ref{tab:bias} present the findings. For the image modality, we have images with positive and negative valence, analogous to positive and negative-sentiment words.
As we can see, the model demonstrates similar bias to the cross-modal case. Similarly, we notice that ``Islam'' and ``Judaism'' attract more negative bias in comparison with ``Christianity'' and ``Buddhism''. In some cases the effect size is slightly different, which can be explained by the fact that the number of samples in the case of textual data is smaller, entailing a greater standard deviation, which in turn alters the effect size.

\subsection{Fusion-based Models}
We next evaluate two fusion-based models. Although these models typically have independent textual and visual encoding modules in their lower layers, their architecture is complemented by a fusion module in higher levels to combine the information in different modalities, enabling them to learn joint embeddings of the visual and textual domains. This has been shown to be essential for more complex tasks such as VQA and NLVR that require more complex reasoning.
The first such model we consider is ALBEF. Similar to CLIP, ALBEF \cite{li2021align} first learns separate visual and textual embeddings using Transformer-based image and text encoders coupled with contrastive loss. However, unlike CLIP, ALBEF further combines these embeddings by adopting an attention-based fusion architecture to model more complex interactions between these modalities, and  directly aims to address several vision--language objectives, including image--text matching and masked language modeling.
This model is pretrained on conceptual captions and SBU captions \cite{sharma2018conceptual,Ordonez:2011:im2text}.
Furthermore, the model is trained using momentum distillation to facilitate learning by adding an auxiliary learning network to stabilize the leaning process. 

ViLT \cite{kim2021vilt} is another recent VLP model that is devised as a more computationally efficient alternative to CLIP and ALBEF. Unlike large and computationally-heavy image and text encoders in CLIP and ALBEF, ViLT utilizes only shallow linear layers to process the sequence of word embeddings and image patches of the text--image input pair. 
Furthermore, in order to enable the model to solve complex vision--language tasks such as VQA, NLVR, and ITM, a Transformer-based architecture is employed on top to capture the complex dynamics between the modalities. This model is trained using a combination of image--text matching, word patch alignment, and masked language modelling objectives. 

With regard to bias assessment, unlike CLIP, fusion-based models do not provide explicit access to separate visual and textual embeddings but rather provide a combined embedding of the pair. As a result, computing the Caliskan distance in Eq.~\ref{eq:bias_def} is not possible. However, interestingly one of the objectives these models optimize for is the image--text matching (ITM) objective. ITM is the problem of estimating the probability that a given image--text pair is a match. This task is directly related to our bias evaluation problem. We can argue that a model is fair if the probability of assigning pleasantness scores is similar across different concepts. In other words, the following should hold for the ITM scores:

\begin{align*}
    P_{\rm ITM}(\boldsymbol{A}|\boldsymbol{X})- P_{\rm ITM}(\boldsymbol{B}|\boldsymbol{X}) \approx\\
    P_{\rm ITM}(\boldsymbol{A}|\boldsymbol{Y})- P_{\rm ITM}(\boldsymbol{B}|\boldsymbol{Y})\ \ 
\end{align*}

Columns 4 and 5 in Table~\ref{tab:bias} include the results for ALBEF and ViLT. The numbers provided are probabilistic differences and are not comparable to the Caliskan scores provided for CLIP. In order to reduce irrelevant noise only the top 15 most significant matches are considered. Again, we see that these models exhibit strong biases favoring Christianity vs.\ Islam and Judaism, matching Christian images to positive words 45\% more than Muslim and 51\% more than Judaism. However, in case of nationality, these models show fewer signs of bias. Furthermore, ViLT and ALBEF show less bias towards physical disabilities compared to mental disability.

\begin{algorithm}[htbp]
\caption{Bias Mitigation Algorithm}\label{alg:cap}
\begin{algorithmic}
\Require

Image Embedding $V^I$, \\ \hspace{32pt} Text Embedding $V^T$,\\ \hspace{32pt} Features to remove $N$,  \\
\hspace{32pt} Classification Labels $L$ 

\State

\State
$X \gets \emptyset$
\State
$\Psi \gets \text{Compute\_Bias}(V^I, V^T)$

\For{$d \gets 1$ \ \ to\ \ $\mathrm{len}(V^I)$}
    \State{$\overline{V^I} \gets V^I\  \textbackslash \ v^I_d$}
    \State{$\overline{V^T} \gets V^T\  \textbackslash \ v^T_d$}
    
    \If{MI($v^I_d,L) < \Theta$}
    \State {$\psi_d \gets \text{Compute\_Bias}(\overline{V^I}, \overline{V^T})$}

    \If{$\psi_d < \Psi$}
        \State {$X\gets X \cup \{(d,\psi_d)\}$}
    \EndIf

     \EndIf

\EndFor

\State {$Z\gets \text{sort}_{\psi_d}(X)[0:N]$} // Dimensions to remove

\State{$X \gets X \ \textbackslash \ Z$}

\State {return $X$} 

\end{algorithmic}
\end{algorithm}

\section{Bias Mitigation Algorithm}

Bias mitigation methods typically fall into one of three categories: 
data augmentation (fair resampling), model adjustment, and embedding post processing algorithms. Each of these alternatives have their own benefits and drawbacks, but a major limitation of the first two is that they require retraining the models. This can be burdensome in many cases. In particular, we often lack access to the dataset, the model's training procedure, or in the case of large pretrained models, retraining may also be computationally infeasible on typical hardware and cost budgets. Post-processing methods, on the other hand, may be invoked as a fast and efficient plug-and-play method to modify learned embeddings without the need for retraining. Since vision--language tasks are complex, VLP models usually have large architectures to be able to capture all the complex dynamics. However, this can cause them to learn redundant or highly correlated features, since they are not optimally compressed. These features are not only computationally wasteful but can also amplify model bias.
Due to the high correlation among some features, we can remove some without affecting performance, while simultaneously reducing bias. In order to identify those features, we directly optimize for the objective in Eq.~\ref{eq:bias_def} by removing features in a greedy manner, pruning $N$ dimensions that cause the largest decrease in bias effect size. However, we only consider the features that exhibit a small mutual information with respect to classification labels. $\Theta$ can be set empirically and this ensures only redundant dimensions are removed.  Algorithm~1 details the steps of our technique. 

\begin{table}[htbp]
\centering
\scalebox{0.85}{
\begin{tabular}{|c|c|c|c|c|c|}
 \hline
Bias & Before & After & Reduction\\
  \hline
Muslim vs Christian & 1.72 & 0.57 & 66\%\\
Jewish vs Christian & 1.69 & 0.75 & 55\%\\
Muslim vs Buddhist & 1.62 & 0.11 & 93\%\\
Jewish vs Buddhist & 1.61 & 0.30 & 82\%\\
Muslim vs Hindu & 1.65 & 0.71 & 57\%\\
\hline
Arab vs American & 1.28 & 0.33 & 74\%\\
Mexican vs American & 1.13 & 0.85  & 26\%\\
Chinese vs American & 0.89 & 0.56 & 38\% \\
\hline
Mental Dis.\ vs No Dis. & 1.48 & 0.49 & 66\% \\
 \hline
LGBTQ\ vs Heterosexual & 1.67 & 0.92 & 45\%\\
  \hline
\end{tabular}
}
\caption{Bias Mitigation Results. Our algorithm is able to significantly reduce bias without substantially affecting performance.} 
\label{tab:debias}
\end{table} 

The results of this debiasing method are presented in Table~\ref{tab:debias}. We removed 54 dimensions (10\% of all dimensions), which leads to up to 93\% bias reduction in some cases. This however only minimally affects the model's classification accuracy. We have tested the accuracy of the model on the MMbias dataset as well as the CIFAR-100 dataset. On MMbias, the accuracy dropped by only 1.1\% , and on CIFAR100 by only 1.3\% from 80.1\% to 78.8\%.
Furthermore, Fig.~\ref{fig:tsne}b shows that even after removing the aforementioned dimensions, the embeddings still remain  well-separable, confirming the redundancy of some of the embedding features.

Regarding the choice of $N$ (number of features removed) in the bias mitigation algorithm,  a larger $N$ will affect the performance of the model more negatively, as previously observed in other dimensionality reduction algorithm. In order to find a reasonable $N$ we can plot the bias reduction as well as performance reduction as a function of $N$. Inspecting this graph allows us to consider the trade-off between greater bias removal and the loss of accuracy, allowing us to choose an $N$ that decreases the bias in a meaningful way while not affecting performance
significantly. 

\section{Conclusion}
Most bias analysis studies focus on gender and racial biases, which is primarily due to a lack of suitable data to consider further important forms of bias. In this study, we have compiled a new multimodal bias assessment dataset called MMBias enabling the study of bias affecting population groups largely neglected in prior studies. Our dataset consists of around 3,500 images and hundreds of phrases covering over 14 minority subgroups. Furthermore, based on a formulation of the bias-fairness problem, we draw on this data to assess the level of bias in several prominent self-supervised multimodal models, including CLIP, ALBEF, and ViLT. Our results show that these models demonstrate meaningful bias towards certain groups. Finally, we introduce a novel bias mitigation technique designed specifically for large pretrained models that can be applied as a post-processing step to reduce bias, and show that it has negligible effects on classification performance as well as data separability. Our data and code is available at \href{https://github.com/sepehrjng92/MMBias}{github.com/sepehrjng92/MMBias}.

\section*{Limitations}
This work seeks to make a contribution towards vision--language models that exhibit less biased behavior. To this end, we provide a large new dataset, new experimental results, and also investigate a bias mitigation method for pre-trained vision--language models. Yet, bias measurement data as well are prone to biases, most notably in the selection of classes and groups, but also with regard to the particular data instances. We envision that MMBias will grow to encompass further groups and additional data in the future, e.g., further ethnic minorities, sexual identities, and gender identities. We also hope that our dataset can serve as a starting point for research on additional natural languages.

Clearly, our bias mitigation algorithm can only mitigate certain fairly overt expressions of bias in vision--language models. Large pre-trained models have millions of parameters that affect the model behavior. As vision--language models necessarily need to rate ties between images and text, they will continue to prefer or disprefer certain associations, leading to remnant biases. Still, we hope that our work will enable the community to pay closer attention to these challenges and work towards models that behave in more equitable ways.

\section*{Ethics Statement}
With our work, we wish to encourage further analysis of bias in machine learning models. To this end, we provide data that enables an assessment of a number of potential manifestations of bias. We acknowledge that the images harbor a multitude of different stereotypes that cannot be taken to be representative of the various groups.
Moreover, we acknowledge that the pairings of classes of people adopted thus far in our work leaves out other groups of people, e.g., further forms of faith and belief, and also further pairings. We view our work as a step towards a more inclusive bias assessment resource that should keep growing in the future.

% \bibliography{anthology,custom}
% \bibliographystyle{acl_natbib}

\appendix

\end{document}